\documentclass[conference]{IEEEtran}
\IEEEoverridecommandlockouts
\usepackage{todonotes}
\usepackage{cite}
\usepackage{amsmath,amssymb,amsfonts}
\usepackage{algorithmic}
\usepackage{graphicx}
\usepackage{textcomp}
\usepackage{xcolor}
\usepackage{booktabs}
\usepackage{hyperref}
\usepackage{multirow}
\def\BibTeX{{\rm B\kern-.05em{\sc i\kern-.025em b}\kern-.08em
    T\kern-.1667em\lower.7ex\hbox{E}\kern-.125emX}}
\begin{document}

\title{UNATE: UNsupervised ATomic Embedding for crystal structures property prediction}

\author{\IEEEauthorblockN{Laura Solà-Garcia}
\IEEEauthorblockA{\textit{Signal Theory and Comms. Dept.} \\
\textit{UPC-Univ. Politècnica de Catalunya}\\
Barcelona, Spain \\
0009-0003-2078-093X}
\and
\IEEEauthorblockN{Àlex Solé}
\IEEEauthorblockA{\textit{Signal Theory and Comms. Dept.} \\
\textit{UPC-Univ. Politècnica de Catalunya}\\
Barcelona, Spain \\
0000-0002-2071-9317}
\and
\IEEEauthorblockN{Javier Ruiz-Hidalgo}
\IEEEauthorblockA{\textit{Signal Theory and Comms. Dept.} \\
\textit{UPC-Univ. Politècnica de Catalunya}\\
Barcelona, Spain \\
0000-0001-6774-685X}

}

\maketitle

\begin{abstract}
Accurately predicting crystal properties is critical for accelerating materials discovery, but it is often limited by scarce labeled data and costly theoretical calculations. To alleviate this, we propose UNATE (Unsupervised Atomic Embedding), a framework that leverages structural information extracted from unlabeled crystal structures. UNATE integrates an unsupervised denoising autoencoder with self-supervised contrastive learning to learn robust atomic representations, which are then used as input features for downstream property prediction. Experimental results show that replacing raw atomic numbers with UNATE-pretrained node embeddings yields a 2.7\% improvement over the full-data baseline. Notably, the benefits become more pronounced in scenarios with limited labeled data, reaching improvements of up to 10\% when only 25\% of the labeled data is used. The code is available at \url{https://github.com/LauraSola/Unsupervised-Atom-Embedding-Gen}.
\end{abstract}


\section{Introduction}
Predicting the properties of crystalline materials is crucial for the discovery of new materials, with applications in electronics, energy storage, and renewable energy~\cite{green,ramprasad,norskov}. Properties such as band gap, formation energy, and elastic constants determine a material’s functionality, but experimental characterization is often slow, costly, and inefficient.

To overcome these limitations, computational models based on theoretical estimations \cite{perdew} have become widely used, though they remain expensive at scale due to the complexity of atomic structures \cite{hautier}. In recent years, machine learning has emerged as a promising alternative, achieving near-theoretical accuracy while enabling large-scale screening ~\cite{CGCNN, materialsProject, louis}.

A major challenge in developing effective machine learning models is the limited availability of high-quality labeled data, which requires expensive theoretical calculations or experimental measurements. In contrast, large public repositories contain crystal structures without associated property labels, motivating the use of techniques that exploit unlabeled data to learn meaningful structural representations~\cite{crysgnn, CrysAtom}.

The contributions for this work are:
\begin{itemize}
    \item A proposed dual-branch self-supervised framework, UNATE, that learns transferable atomic embeddings from unlabeled crystal graphs, and transfer them by replacing raw atomic numbers without modifying downstream architectures.
    \item We demonstrate consistent improvements in supervised band gap prediction, including improved label efficiency (up to 10\% MAE improvement with 25\% labeled data, and +2.7\% with 100\% labeled data training).
    \item We provide qualitative evidence via t-SNE visualizations that UNATE embeddings recover meaningful chemical-space structure directly from data, without encoding explicit chemistry theory.
\end{itemize}

\section{Related Work}

Graph Neural Networks (GNNs) are widely used for molecular and crystal property prediction due to their ability to model relational structures. Early models such as CGCNN~\cite{CGCNN} represented crystals as periodic graphs with periodic boundary conditions (PBC). More recent architectures improve geometric reasoning and long-range interaction modeling, including Matformer~\cite{Matformer}, which integrates lattice information into a transformer framework; PotNet~\cite{PotNet}, which captures local and global interactions using physically motivated potentials; iComformer/eComformer~\cite{conformers}, which enforces rotation invariance through refined unit-cell representations or rotation equivariance via tensor products; CartNet~\cite{CartNet}, which achieves soft rotation-equivariance via rotational data augmentation, improving the speed and accuracy; and PRISM~\cite{prism}, which explicitly integrates multiscale representations and periodic feature encoding through a mixture of expert modules.

Because labeled crystal datasets are scarce due to the high cost of experiments and simulations, recent work has focused on pretraining GNNs on unlabeled crystal structures to learn transferable representations.
DualSSL~\cite{dualSSL} proposes a multi-task self-supervised framework combining atom masking, contrastive learning between perturbed views, and micro-property prediction (like atomic stiffness) to capture both local and global structure, followed by supervised fine tuning of the pretrained backbone.
CrysGNN~\cite{crysgnn} trains a teacher model using both self-supervised (like node reconstruction) and supervised objectives (like space-group classification), and transfers knowledge to downstream models via distillation, enabling architecture-independent reuse of learned representations by aligning intermediate features between the teacher and student models.
Finally, CrysAtom~\cite{CrysAtom} learns transferable atom-level embeddings using an autoencoder with a CGCNN backbone that reconstructs node and edge attributes and applies a Barlow Twins loss~\cite{Barlow} to promote invariant and non-redundant representations. These embeddings are then reused in downstream models as input node features.
Previous pretraining approaches are often evaluated on relatively simple backbones, where gains are easier to obtain but tend to diminish as model capacity increases. In contrast, our method yields consistent improvements even with high-performing backbones. In addition, by transferring knowledge through pretrained embeddings rather than model weights, the approach remains model- and task-agnostic, enabling broad applicability across downstream architectures and property prediction tasks.

\section{Methodology}
\begin{figure}[h!!]
    \centering
    \includegraphics[width=\linewidth]{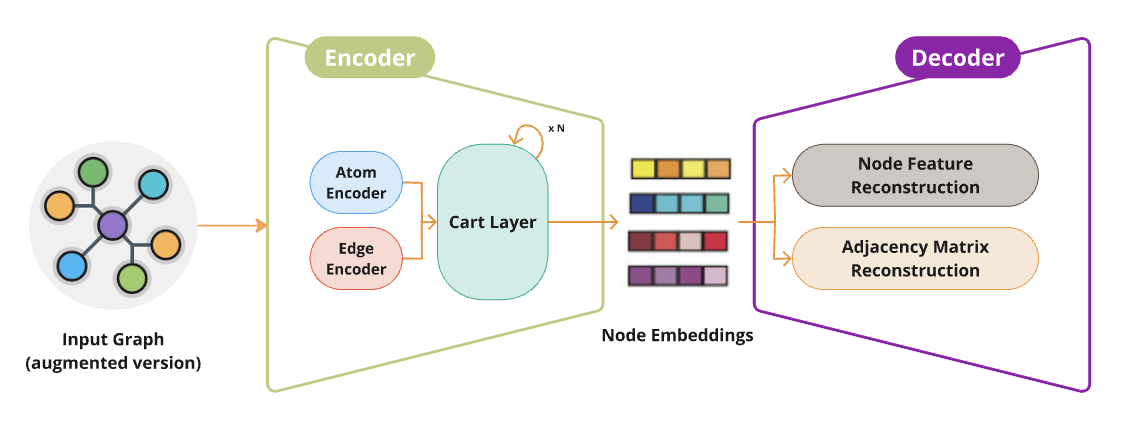}
    \caption{Unsupervised learning branch. CartNet encodes a masked, edge-dropped graph. The decoders reconstruct atomic numbers and the periodic-aware adjacency matrix.}
    \label{fig:unsup_branch}
\end{figure}

This work aims to enhance crystal property prediction by learning transferable representations from unlabeled crystal graphs that can be used to improve downstream supervised tasks. We propose a dual-branch pretraining framework combining an unsupervised denoising autoencoder and a self-supervised contrastive learning objective. The two branches are trained in parallel, encouraging the encoder to capture both local structural information and global semantic consistency, as illustrated for the unsupervised branch in Fig.~\ref{fig:unsup_branch}.

\subsection*{Graph representation and augmentations} \label{section:aug}
Each crystal is represented as a periodic graph $G = (V, E)$, where nodes $i \in V$ correspond to atoms and edges $(i,j) \in E$ denote atomic interactions under periodic boundary conditions. Node features are atomic numbers $Z_i$, and edge features include interatomic distances and direction vectors. These features are then mapped to high-dimensional embeddings, \(\mathbf{h}_i \in \mathbb{R}^{d}\) for nodes and \(\mathbf{h}_{ij} \in \mathbb{R}^{d_e}\) for edges, encoding chemical and structural information for GNN layers to propagate and aggregate information across the graph. 

Lastly, to improve embedding robustness, we apply two augmentations to the graphs: node masking, which randomly hides a subset of node features to encourage inference from context, and edge dropping, which randomly removes edges to promote learning from remaining interactions.

\subsection{Unsupervised Learning Branch: Denoising Autoencoder}
The unsupervised branch follows a denoising autoencoder architecture. The encoder receives an augmented graph (as described in the section~\ref{section:aug}) and reconstructs the original graph, learning robust embeddings \( \mathbf{h}_i \) that capture local chemical environments and structural context. This denoising objective encourages inference from context rather than memorization, improving downstream performance when combined with supervised fine tuning (Section~\ref{section:ablation}).

For the encoder, we adopt CartNet~\cite{CartNet} due to its strong and lightweight performance relative to other crystal graph architectures. The CartNet encoder consists of an atom encoder, an edge encoder, and several CartLayers. It processes an augmented graph \( G = (V, E) \), with masked nodes and dropped edges, and outputs node embeddings \( \mathbf{h}_i \), which are then passed to the two decoders.

\subsubsection{Node Feature Reconstruction (decoder 1)}

The node embeddings \( \mathbf{h}_i \in \mathbb{R}^d \) are first L2-normalized and then passed through a linear layer that projects each embedding vector into a 118-dimensional space, corresponding to the number of atomic element classes. The resulting logits are then passed through a softmax function to obtain the predicted probability distribution \(\hat{\mathbf{y}}_i\):

\begin{equation}
\hat{\mathbf{y}}_i =
\text{softmax}\!\left(
\mathbf{W}_{\mathrm{node}} \cdot \frac{\mathbf{h}_i}{\|\mathbf{h}_i\|_2}
+ \mathbf{b}_{\mathrm{node}}
\right),
\quad
\hat{\mathbf{y}}_i \in \mathbb{R}^{N_Z},
\end{equation}

for each node \(i \in V\), where \(\mathbf{W}_{\mathrm{node}} \in \mathbb{R}^{N_Z \times d}\) and \(\mathbf{b}_{\mathrm{node}} \in \mathbb{R}^{N_Z}\) are learnable parameters, and \(N_Z\) denotes the number of possible atomic numbers. The model is trained by minimizing the negative log-likelihood loss \(\mathcal{L}_{\text{node}}\), aligning the predicted atomic number distribution with the ground-truth labels.

\subsubsection{Adjacency Matrix Reconstruction (decoder 2)}

This decoder reconstructs the graph connectivity by predicting the number of edges originally present between each node pair after random edge removal. Due to Periodic Boundary Conditions (PBC), multiple edges can exist between the same pair of atoms. Edge reconstruction is therefore formulated as a six-class classification problem, corresponding to edge multiplicities: 0, 1, 2, 3, 4, or 5+.

For each node pair \((i,j)\), the decoder first computes a bilinear interaction vector from the normalized node embeddings:

\begin{equation}
\mathbf{s}_{ij} = \mathbf{h}_i^\top \mathbf{W}_b \mathbf{h}_j + \mathbf{b}_b,
\end{equation}

where \(\mathbf{W}_b \in \mathbb{R}^{d \times 6 \times d}\) and \(\mathbf{b}_b \in \mathbb{R}^{6}\) are the weights and bias of the bilinear layer. This interaction vector $\mathbf{s}_{ij}$ is then projected through a linear layer to produce logits, which are normalized with a softmax to obtain edge probabilities ($\hat{\mathbf{y}}_{ij}$):

\begin{equation}
\hat{\mathbf{y}}_{ij} = \text{softmax}(\mathbf{W}_a \mathbf{s}_{ij} + \mathbf{b}_a),
\end{equation}

where \(\mathbf{W}_a \in \mathbb{R}^{6 \times 6}\) and \(\mathbf{b}_a \in \mathbb{R}^{6}\) are the trainable parameters of the linear layer.

The model is trained using a weighted cross-entropy loss, denoted as $\mathcal{L}_{\text{adj}}$, assigning lower weight to the zero-edge class to mitigate class imbalance caused by the large number of non-connected node pairs.

\subsection{Self-supervised Learning Branch: Contrastive Objective}

In parallel to the denoising autoencoder, we use a contrastive self-supervised branch to encourage globally consistent and semantically structured embeddings. Instead of reconstructing features, this branch drives the encoder to organize the embedding space so that similar crystal structures are embedded closer together, improving generalization and robustness. 

Graph-level embeddings are obtained by mean-pooling node embeddings produced by the encoder and passing them through a two-layer MLP with SiLU activation:

\begin{equation}
\mathbf{z} = 
\mathbf{W}_2 \cdot 
\text{SiLU}\!\left(
\mathbf{W}_1 
\left( \frac{1}{|V|} \sum_{i \in V} \mathbf{h}_i \right)
+ \mathbf{b}_1
\right)
+ \mathbf{b}_2.
\end{equation}

For each graph, we generate two random augmented views via node masking and edge dropping. Let \( (\mathbf{z}_i, \mathbf{z}_j) \) be a positive pair (embeddings originating from the same graph), and all other embeddings in the batch act as negatives.

For each anchor embedding \( \mathbf{z}_i \), the contrastive InfoNCE loss~\cite{infoNCE} encourages similarity to its positive counterpart \( \mathbf{z}_j \) while discouraging similarity with all negatives:

\begin{equation}
\mathcal{L}_{\text{InfoNCE}}^{(i)} = 
-\log\!\left(
\mathrm{Softmax}_{k \neq i}
\left(
\frac{1}{\tau} \cdot \text{sim}(\mathbf{z}_i, \mathbf{z}_k)
\right)_j
\right),
\label{eq:contr_loss}
\end{equation}

with temperature \( \tau \). The final loss, denoted as $\mathcal{L}_{\text{InfoNCE}}$, is the average over all \( 2N \) augmented embeddings.

This objective enforces invariance to graph perturbations while preserving discriminative structure, complementing the reconstruction branch and enhancing learned representations.

\subsection{Total Pretraining Objective}

\begin{figure}[h!!]
    \centering
    \includegraphics[width=1\linewidth]{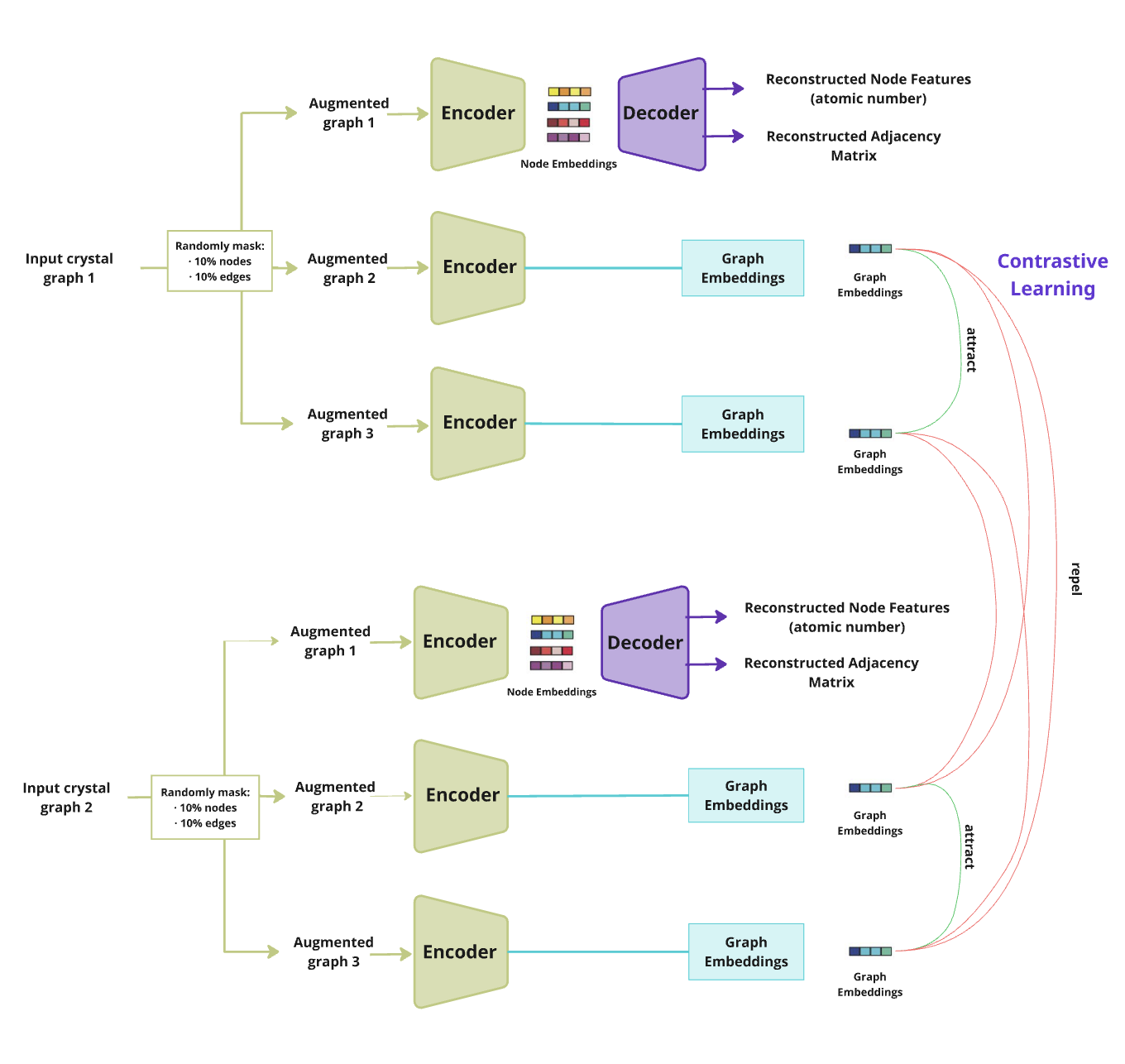}
    \caption{For each crystal graph, we generate multiple augmented views by randomly masking nodes and edges. A shared encoder produces node embeddings that a decoder uses to reconstruct the masked atomic features and adjacency relations (denoising objective). In parallel, graph-level embeddings from each augmented view are optimized with a contrastive loss that pulls together embeddings of the same crystal and pushes apart those from different crystals.}
    \label{fig:framework}
\end{figure}

Fig.~\ref{fig:framework} summarizes the overall training setup and how the loss terms in Eq.~\eqref{eq:total_loss} are obtained. The final objective combines the reconstruction and contrastive losses:

\begin{equation}
\mathcal{L}_{\text{total}} = \alpha \, \mathcal{L}_{\text{node}} + \beta \, \mathcal{L}_{\text{adj}} + \gamma \, \mathcal{L}_{\text{InfoNCE}}
\label{eq:total_loss}
\end{equation}

where \( \alpha, \beta, \gamma \) control the contribution of each loss term.

\subsection{Knowledge Transfer via Pretrained Atomic Embeddings}

Knowledge learned during pretraining is transferred to downstream crystal property prediction tasks through pretrained atomic embeddings. Specifically, raw atomic numbers are replaced by fixed, element-wise embeddings learned from unlabeled crystal structures, which then serve as node features for the downstream model.

To construct these embeddings, the pretrained encoder is frozen and applied to the training dataset to extract node representations. The embeddings corresponding to each atomic number are aggregated by computing their mean across all occurrences, resulting in a single, stable representation per element. The final embeddings are L2-normalized to ensure consistent scaling.

In downstream models, the Atom Encoder (component of CartNet~\cite{CartNet}) is replaced by a lightweight linear projection that adapts the pretrained embeddings before they are processed by the message-passing layers. Apart from this modification, the downstream architecture is trained from scratch, without reusing pretrained weights.

\section{Results}

We first describe the datasets and computational resources used for evaluation, followed by a quantitative analysis of the proposed pretraining framework. All results are reported as the average over four independent runs to account for the variability inherent to graph neural network training.

\subsection*{Datasets}
We use two datasets from the Materials Project~\cite{materialsProject}. The unlabeled dataset contains $\sim$139,000 crystal structures represented as crystal graphs, used for the generation of the embeddings. The labeled dataset (2018.6.1 release) comprises 69,239 inorganic crystals with various property labels. Band gap was selected as the target property because it is available for a much larger number of materials than bulk or shear modulus, which are limited to $\sim$5,500 crystals, and has been commonly used to evaluate representation learning methods in prior work~\cite{CrysAtom}.

\subsection*{Training details}

All experiments were executed on a \href{https://tsc.upc.edu/en/it-services/computing-services/copy_of_gpi-inventory}{high-performance computing cluster} using a single GPU per run, with 11 to 48 GB of GPU memory and up to 6 CPU cores allocated per experiment, depending on the computational requirements.

CartNet was employed as the backbone. During training for embedding generation, the Adam optimizer~\cite{adam} was used with a learning rate of $3\cdot10^{-2}$ and batch size of 128 for 100 epochs. The parameters for the final loss (Eq.~\ref{eq:total_loss}) were set to $\alpha = 225$, $\beta = 4$, and $\gamma = 3$. Hyperparameter tuning was performed using grid search.

\subsection{Results and Impact of Labeled Dataset Size}

From Table~\ref{tab:main_result}, we compare supervised band gap prediction performance using 128-dim embeddings across different pretraining strategies using CartNet as the backbone. The results show that UNATE embeddings improve the MAE by 2.7\% with respect to the baseline, whereas CrysAtom embeddings do not yield gains over the baseline. 

\begin{table}[htbp]
\caption{Supervised band gap prediction (128-dim embeddings). MAE ($\pm$ std.\ dev.; $\times 10^{-3}$ (Improv. \%)}
\centering
\begin{tabular}{l c}
\toprule
\textbf{Method} & \textbf{Bandgap} \\
\midrule
CartNet & $194.02 \pm 3.05$ \\
w/ CrysAtom~\cite{CrysAtom} & $195.89 \pm 2.61$ \,(-0.6\%) \\
w/ UNATE & $\mathbf{188.78 \pm 2.19}$ \,(\textbf{+2.7\%}) \\
\bottomrule
\end{tabular}
\label{tab:main_result}
\end{table}

As this approach is aimed at reducing the dependence on labeled  data, we progressively reduced the size of the dataset and measured the corresponding improvement provided by pretraining. Starting from the full set of $\sim$69K crystals, we also considered 50\% and 25\% subsets, while keeping the pretraining fixed on the full unlabeled dataset.

\begin{table}[htbp]
\caption{Impact of labeled dataset size on MAE ($\pm$ std.\ dev.; $\times 10^{-3}$) for multiple embedding dimensions using CartNet as backbone.}
\centering
\resizebox{\linewidth}{!}{%
\begin{tabular}{c c c c}
\toprule
\textbf{Dim} & \textbf{\% Labeled Data} & \textbf{\textit{Baseline}} & \textbf{\textit{Pretrained (Improv.\ \%)}} \\
\midrule
\multirow{3}{*}{64}  & 100\% & $202.47 \pm 2.23$ & $197.77 \pm 1.51$ (+2.3\%) \\
                     & 50\%  & $241.60 \pm 2.02$ & $235.22 \pm 3.23$ (+2.6\%) \\
                     & 25\%  & $291.67 \pm 4.65$ & $274.58 \pm 5.97$ (+5.9\%) \\
\midrule
\multirow{3}{*}{128} & 100\% & $194.02 \pm 2.62$ & $188.78 \pm 2.19$ (+2.7\%) \\
                     & 50\%  & $240.55 \pm 8.19$ & $220.91 \pm 4.27$ (+8.2\%) \\
                     & 25\%  & $289.61 \pm 10.24$ & $260.63 \pm 7.00$ (+10.0\%) \\
\bottomrule
\end{tabular}%
}
\label{tab:mae_results_comparison}
\end{table}



As shown in Table~\ref{tab:mae_results_comparison}, the benefit of pretraining grows as labeled data becomes scarce. With 64-dimensional embeddings, the relative improvement increases from 2.3\% (full data) to 5.9\% with only 25\% of the labeled samples. For 128-dimensional embeddings, the gains rise from 2.7\% (full data) to 10\% at 25\% of labeled data. These results highlight the effectiveness of pretraining, particularly in low-data regimes, where pretrained node embeddings provide chemical and structural information beyond atomic numbers.

\subsection{Ablation Study} \label{section:ablation}
Table~\ref{tab:mae_results_comparison} shows that 128-dimensional embeddings outperform 64-dimensional embeddings for both the baseline and pretrained models (larger embedding sizes could not be further explored due to GPU memory limitations). Pretraining consistently improves performance across both embedding sizes. Although the main experiments use 128-dimensional embeddings, the following ablation studies are conducted with 64-dimensional embeddings to reduce computational cost. Despite the lower dimensionality, the 64-dimensional setting preserves the same overall performance trends and is therefore sufficient for analyzing the contribution of individual components of the framework.

Using the 64-dimensional embeddings, we analyzed the contribution of individual components of the proposed framework. Three aspects were considered: the effect of noise injection in the denoising autoencoder, the choice of self-supervised objective, and the strategy used to transfer pretrained knowledge. Results are summarized in Table~\ref{tab:ablation}. 

\begin{table}[htbp]
\caption{Ablation study covering noise injection, self-supervised method, and knowledge-transfer strategy using CartNet as backbone. MAE ($\pm$ std.\ dev.; $\times 10^{-3}$) .}
\centering
\begin{tabular}{l c}
\toprule
\textbf{Experiment / Setting} & \textbf{MAE} \\
\midrule
CartNet (baseline - no pretraining) & $202.47 \pm 2.30$ \\
\midrule
\multicolumn{2}{l}{\textbf{1. Autoencoder Noise}} \\
\midrule
No Noise & $201.77 \pm 2.65$ (+0.35\%) \\
Noise Added (UNATE) & \textbf{$197.77 \pm 1.51$ (+2.32\%)} \\
\midrule
\multicolumn{2}{l}{\textbf{2. Self-Supervised Method}} \\
\midrule
Barlow Twins~\cite{Barlow} & $205.25 \pm 2.61$ (-1.37\%) \\
Contrastive Learning (UNATE) & $\mathbf{197.77 \pm 1.51}$ (+2.32\%) \\
Deep Graph Infomax~\cite{DGI} & $199.21 \pm 2.18$ (+1.61\%) \\
\midrule
\multicolumn{2}{l}{\textbf{3. Knowledge Transfer Strategy}} \\
\midrule
Weight Initialization (transfer learning) & $417.89 \pm 4.01$ (-106.4\%) \\
Node Embeddings (UNATE) & $\mathbf{197.77 \pm 1.51}$ (+2.32\%) \\
Combined (weight init.\ + node embd.) & $511.89 \pm 5.67$ (-152.9\%) \\
\bottomrule
\end{tabular}%
\label{tab:ablation}
\end{table}

From these ablation studies, three main observations emerge. Adding noise via node masking and edge dropping during pretraining consistently improves downstream performance, confirming the regularizing effect of the denoising autoencoder. Among the self-supervised objectives, contrastive learning yields the best results, suggesting that distinguishing between different structural contexts while maintaining invariance across augmented views leads to more informative representations. Finally, transferring pretrained atomic embeddings enhances performance compared to the baseline, whereas direct weight initialization from the pretrained encoder severely degrades performance, likely due to a mismatch with the downstream prediction task.

\subsection{Visualizing the Embeddings}

\begin{figure}[h!]
    \centering
    \includegraphics[width=\linewidth]{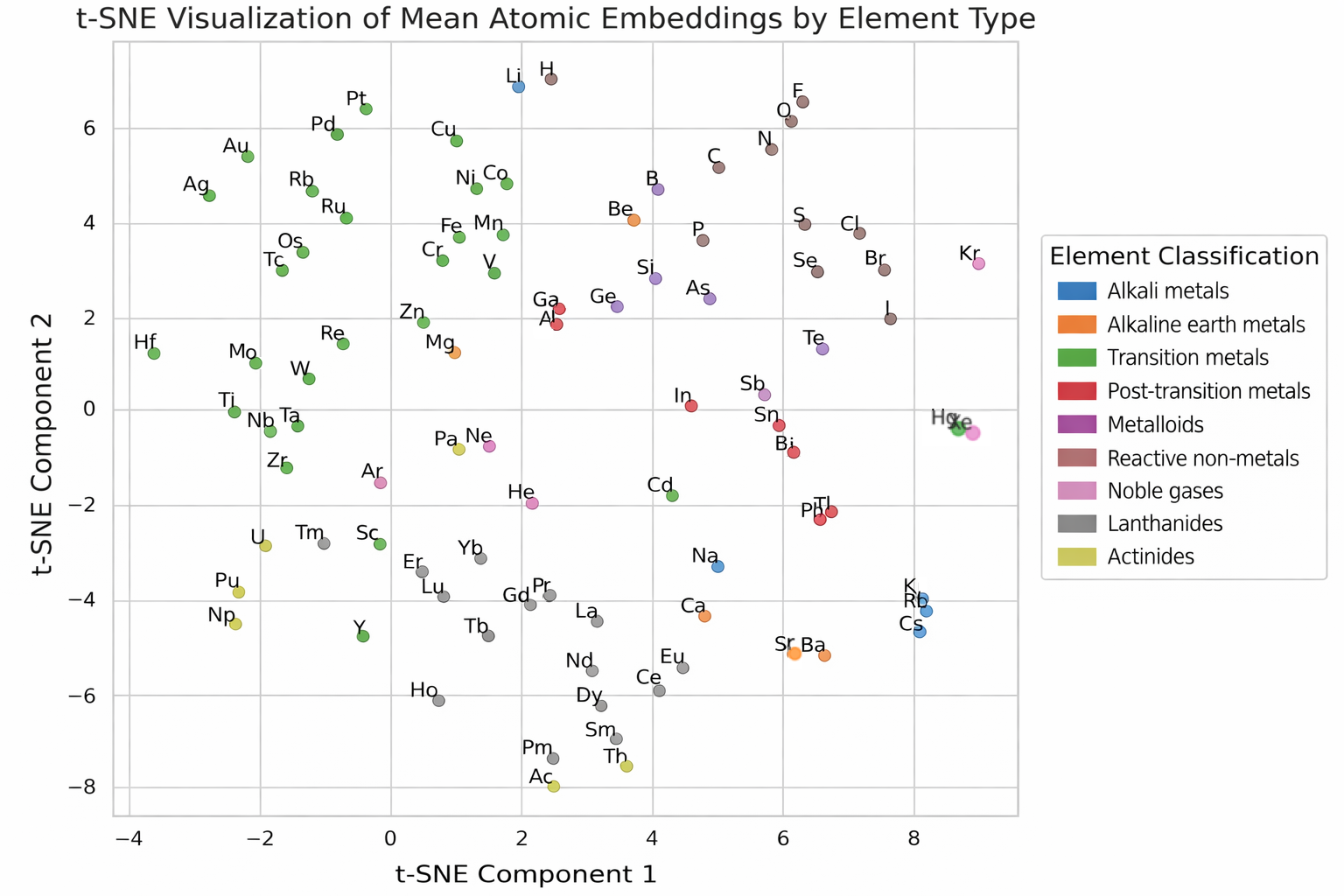}
    \caption{t-SNE visualization of the 128-dimensional atomic embeddings (first and second components). Each point corresponds to an element (mean embedding by atomic number), colored by its chemical category.}
    \label{fig:tsne_mean}
\end{figure}

Finally, to assess the quality and interpretability of the learned representations, we visualize the atomic embeddings using t-SNE~\cite{tsne}. Figure~\ref{fig:tsne_mean} shows a t-SNE projection of the pretrained node embeddings (averaged by atomic number), later used as input for the downstream prediction task. Clear clusters emerge according to chemical families: transition metals (green) group in the upper left, lanthanides (gray) in the lower center, and reactive non-metals toward the upper right, etc. This demonstrates that the embeddings capture chemically meaningful information, reflecting both structural and elemental properties, which leads to improved downstream property prediction.

\section{Conclusions}
We have presented a dual-branch pretraining framework for crystal property prediction, combining a denoising autoencoder and a contrastive self-supervised branch. Using pretrained node embeddings as input features consistently improved bandgap prediction performance. Specifically, improvements of 2.3\% and 2.7\% were observed for 64- and 128-dimensional embeddings, respectively, with larger gains (up to 10\%) in low-data regimes, highlighting the benefits of pretraining when labeled data is limited.

Our analysis shows that contrastive learning outperforms other self-supervised approaches such as Barlow Twins and Deep Graph Infomax. Furthermore, t-SNE projections of the learned embeddings reveal meaningful chemical structure, with atoms from the same group in the periodic table clustering together, despite training being fully unsupervised.

These results demonstrate that pretraining on large unlabeled crystal datasets is an effective strategy to reduce the computational cost of creating labeled data and enhance downstream property prediction. Future work could focus on evaluating the proposed framework on additional material properties, such as formation energy and bulk modulus, to further assess its generalizability beyond bandgap prediction. In addition, exploring alternative self-supervised learning strategies, including triplet loss and contrastive learning with hard negative sampling, as well as different graph augmentation techniques, such as edge rewiring and node feature shuffling, may lead to richer and more informative node representations.

\section*{Acknowledgment}

This work has been supported by the research project PID2024-161868OB-I00 [C3DRUM] funded by MCIN/AEI/10.13039/501100011033 and FEDER, EU.

\bibliographystyle{plain} 
\bibliography{bibliography}

\end{document}